\documentclass{article}
\usepackage{spconf,amsmath,graphicx}

\usepackage{algorithm}
\usepackage[noend]{algpseudocode}
\usepackage{amssymb}
\usepackage{amsmath}
\usepackage{booktabs}
\usepackage{color}
\usepackage{boldline}
\def\*#1{\mathbf{#1}}
\usepackage[T1]{fontenc}
\usepackage[font=small,labelfont=bf,tableposition=top]{caption}
\DeclareCaptionLabelFormat{andtable}{#1~#2  \&  \tablename~\thetable}
\usepackage[export]{adjustbox}
\usepackage{pifont}
\newcommand{\cmark}{\ding{52}}%
\usepackage[outline]{contour}
\contourlength{0.01pt}
\contournumber{10}%
\usepackage{wrapfig}
\usepackage{subfig}
\def\c{\checkmark}
\usepackage{arydshln}
\usepackage{url}

\title{EFFICIENT DIALOG POLICY LEARNING VIA POSITIVE MEMORY RETENTION}
\name{Rui Zhao, Volker Tresp}
\address{Ludwig Maximilian University, Oettingenstr. 67, 80538 Munich, Germany \and Siemens AG, Corporate Technology, Otto-Hahn-Ring 6, 81739 Munich, Germany\\}
\begin{document}
%
\maketitle
\begin{abstract}
This paper is concerned with the training of recurrent neural networks as goal-oriented dialog agents using reinforcement learning.
Training such agents with policy gradients typically requires a large amount of samples.
However, the collection of the required data in form of conversations between chat-bots and human agents is time-consuming and expensive. 
To mitigate this problem, we describe an efficient policy gradient method using positive memory retention, which significantly increases the sample-efficiency. 
We show that our method is 10 times more sample-efficient than policy gradients in extensive experiments on a new synthetic number guessing game.
Moreover, in a real-word visual object discovery game, the proposed method is twice as sample-efficient as policy gradients and shows state-of-the-art performance. 
\end{abstract}
\begin{keywords}
Goal-Oriented Dialog System, Deep Reinforcement Learning, Recurrent Neural Network
\end{keywords}
\section{Introduction}
\label{sec:intro}

In recent years, advances in Deep Learning (DL) \cite{goodfellow2016deep, zhao2017two} and Reinforcement Learning (RL) \cite{sutton1998reinforcement} have led to tremendous progress across many areas of natural language processing (NLP), gameplay \cite{sutskever2014sequence,silver2016mastering,lecun2015deep,lewis2017deal,vinyals2015show}, and robotics \cite{ng2006autonomous, peters2008reinforcement, levine2016end, zhao2018energy}. This progress, in turn, generated an emerging research area, the learning of goal-oriented dialogs \cite{bordes2016learning}.
This research involves agents that conduct a multi-turn dialogue to achieve some task-specific goal, such as locating a specific object in a group of objects \cite{end_to_end_gw}, inferring which image the user is thinking about \cite{visdial_rl}, and providing customer services and restaurant reservations \cite{bordes2016learning}. All these tasks require that the agent possesses the ability to conduct a multi-round dialog and to track the inter-dependence of each question-answer pair. Eventually, the agent learns an optimal policy through trial-and-error. The reward signal of each trail is delayed, and is only available at the end of the dialog. Also the reward signal is very sparse compared with a vocabulary size that often exceeds several thousands. Due to these challenges, in practice, policy gradient methods \cite{williams1992simple} perform more favorably than Q-learning methods \cite{sutton2018reinforcement}. 

\begin{figure}
\begin{minipage}{0.25\linewidth}
\includegraphics[width=\textwidth]{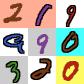}
\end{minipage}
\hspace{0.1cm}
\begin{minipage}{0.65\linewidth}
  \begin{tabular}{p{0.1cm} p{5cm}}
    \# & Question  \hfill{Answer} \\ \hline
    1 & Is it 2 in the image?  \hfill{No} \\
    2 & Is it in a yellow background?  \hfill{No} \\
    3 & Is it 9 in the image?  \hfill{Yes} \\
    4 & Is it in a white background?  \hfill{Yes} \\
    5 & Is it a stroke style digit?  \hfill{Yes} \\
    6 & Is it a digit in blue?  \hfill{No} \\ 
      & Guess: row 1 column 3  \hfill{\cmark} \\ 
  \end{tabular}
\end{minipage}
\caption{\textbf{MNIST GuessNumber dataset example:} Each sample consists of an image (left), a set of sequential questions with answers (right), and a target digit. The goal of this game is to find out the target digit by a multi-round question-answering.} 
\label{fig:MNIST-GuessNumber-example-0}
\end{figure}

Consider a simple goal-oriented dialog example from our synthetic dataset in Figure \ref{fig:MNIST-GuessNumber-example-0}. 
We initialize three roles in this number guessing game, i.e.\ a \textit{questioner}, an \textit{answerer}, and a \textit{guesser}.
The questioner and the guesser try to infer which number the answerer is thinking about. First, the questioner asks questions  about the target digit given the image, such as the color of the digit, the background color, the style of the digit, and also the number itself. Then the answerer responses with a yes/no answer. The questioner needs to reason based on the history dialog and keeps querying with meaningful questions. 
At the end, when the maximum number of questions is reached, the guesser analyzes the whole conversation along with the image, and takes a guess. If the guess is correct, then the task is completed successfully, and the questioner gets a positive reward signal. Otherwise the task is counted as a failure, and the questioner gets a non-positive reward signal.

The training of chat-bots using on-policy policy gradient methods requires numerous training samples. When the samples are generated through human-machine-interaction, e.g.\ by using the Amazon Mechanical Turk or in real-world applications, the collection of the data is time-consuming and expensive \cite{visdial_eval}. Hence, sample-efficiency receives increasingly more attention in dialog policy learning.
We improve sample-efficiency using a novel on-off-policy policy gradient method relying on a biologically inspired mechanism \cite{gruber2016post}, termed positive memory retention. This mechanism employs a bounded importance weight proposal on past positive trajectories, i.e.\ the behavior policy \cite{degris2012off}, to train the target policy network. The retention stops automatically when no further improvement occurs in a predefined number of iterations. The bounded importance weight proposal tackles the problem of high variance in importance sampling. To reduce variance even further, we use recent behavior policies to update the probability in the memory buffer. An early-stopping mechanism within each epoch provides a trade-off between the sample-efficiency and the computational cost. 

\textbf{Contributions:} (1) We  introduce positive memory retention for efficient dialog policy learning, which uses bounded importance sampling, probability updating, and adaptively adjusts the retention times via early stopping within epochs. (2) We perform a comprehensive study about the performance of our method for goal-oriented dialog tasks using a new synthetic number guessing game and verify the high sample-efficiency of our algorithm. 
(3) The proposed model is also tested on a real-world benchmark GuessWhat?!\ game \cite{guesswhat_game} and shows state-of-the-art performance, and an increased sample-efficiency by a factor of two.

\section{Background}
\label{sec:background}
This section introduces recurrent language models, Markov decision process, policy gradient, and importance sampling.

\textbf{Recurrent Language Models:}
The goal of a recurrent neural network (RNN) based language model in NLP is to produce an output sequence $y=[a_1,a_2,...,a_T]$ given a context $x$ as input \cite{mikolov2010recurrent}. Here $a_i \in \mathcal{A}$ where $\mathcal{A}$ is the word vocabulary. 
For each step, the recurrent unit processes the previous word along with the context, and outputs a new word. 
At each time step $t$, the state $s_t$ is the context input $x$ and the words $y_{t-1}=[a_1, ... , a_{t-1}]$ produced by the RNN so far, i.e. $s_t=(x, y_{t-1})$. We sample the next word $a_t$ from this probability distribution $\pi(\cdot|s_t)$, then update our state $s_{t+1}=(x, y_t)$ where $y_t=[y_{t-1}, a_t]$, and repeat in a similar fashion. 


\textbf{Markov Decision Process and Policy Gradient:}
\begin{figure}
\centering
\includegraphics[width=0.35\textwidth]{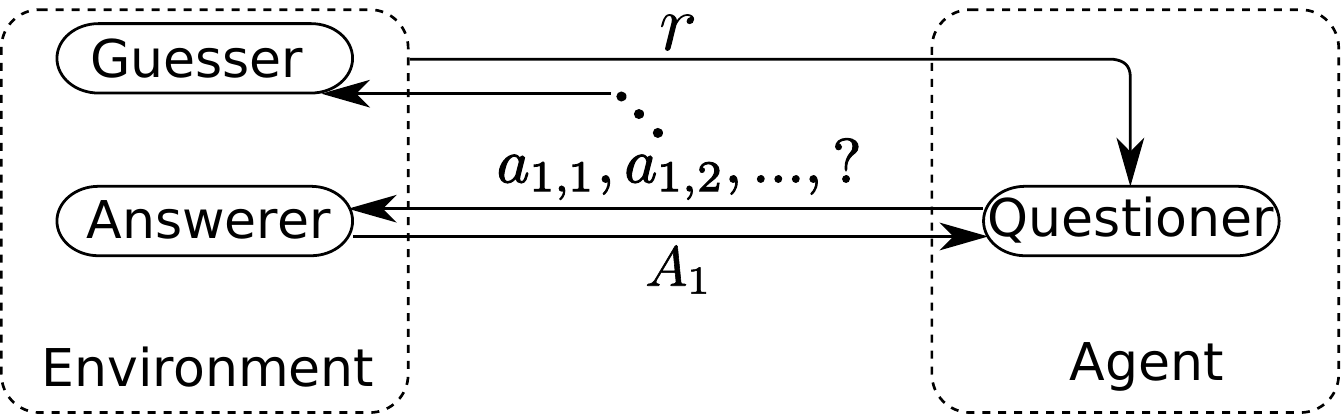}
\caption{\textbf{GuessNumber Gameplay:} The gameplay of the guessing game involves three plays, a questioner, an answerer, and a guesser. In our settings, we first pretrain all three models, then post-train only the questioner using RL. For each gameplay, the questioner first asks a question word by word, $a_{1,1}$, $a_{1, 2}$, ..., ?, then the answerer responses with an answer, $A_1$. This question-answering repeats for a predefined number of rounds. Finally, the guesser reads in the dialog and makes a guess. If the guess is correct, then the questioner receives a reward $r=1$, otherwise, $r=0$.} 
\label{fig:gameplay}
\end{figure}
We formalize a simplified Markov decision process (MDP) to our setting. In the MDP, an agent takes an action $a$ in a state $s$ and transitions to a new state $s'$. A trajectory $\tau$ refers to a sequence of transitions 
until the agent enters a terminal state 
where it receives a reward from the environment. 

In our guessing games, a trajectory $\tau$ is ($x$, $a_{1,1}$, $a_{1, 2}$, ..., ?, $A_1$, $a_{2,1}$, $a_{2,2}$, ..., ?, $A_2$, ..., $G$, $r$), where $x$ is the context (image); $a_i$ is the word sequence in question $i$; ?\ is the question mark that only occurs at the end of each question; $A_i$ is the answer to question $i$; $G$ is the output of the guesser; $r$ is the reward for being correct or incorrect. See also Figure \ref{fig:gameplay}.
 
Formally, the simplified MDP is a triple of $(\mathcal{S},\mathcal{A},R)$ where $\mathcal{S}$, $\mathcal{A}$, and $R$ represent a set of states, actions, and rewards, respectively. 
A policy $\pi$ is a function that chooses an action at a given state, e.g.\ $\pi : \mathcal{A} \times \mathcal{S} \rightarrow \mathbb{R}$, where $\pi(a | s)$ refers to the probability of executing action $a$ at the state $s$. 
When we sample an action $a_t \sim \pi(\cdot | s_t)$, we transition into state $(x,[y_{t-1}, a_t])$. 
We overload notation and let $R(\tau) = \sum_{t=1}^{T}R(s_t)$ be the accumulated reward of a trajectory $\tau$. We define that $R(\tau) = 1$, if the task is a success; $R(\tau) = 0$, if the task is a failure. 
The policy network $\pi$ is a recurrent language model, parametrized by a vector $\theta \in \mathbb{R}^n$, i.e.\ $\pi_{\theta}$. The expected return of a policy $\pi_{\theta}$ is:
$$J(\theta) = \mathbb{E}[R(\tau)|\pi_{\theta}].$$
Our goal is to learn $\theta$ to maximize the expectation of the return $J(\theta)$. The objective function can be optimized with an on-policy policy gradient method, known as REINFORCE \cite{williams1992simple}. The gradient is calculated as: 
$$\nabla_{\theta} J(\theta) = \nabla_{\theta} \mathbb{E} \left[(R^{}(\tau) - b) \mathrm{log} \pi_{\theta} (a_t|s_t) \right]$$
where $b$ is an optional baseline function used to reduce the variance of the gradient estimate \cite{williams1992simple}.

\textbf{Importance Sampling:}
Importance sampling (IS) is a general technique to estimate an integral $\int f(x) p(x) dx$ of a function $f(x)$, with distribution $p(x)$ \cite{mcbook}.
IS samples from an appropriate proposal distribution $q(x)$, and then uses the samples to estimate the integral:
$$I = \mathbb{E}[f] = \int f(x) \frac{p(x)}{q(x)} q(x) dx \approx \frac{1}{N} \sum_{n=1}^{N} \omega_n f(x^{(n)}) = \hat{I},$$
where $\omega_{n} = {p(x^{(n)})}/{q(x^{(n)})}$ are the importance weights.
The goal is  to minimize the variance of the estimate $\hat{I}$, which is proportional to
$\mathrm{var}_{q}[f(x) \omega (x))] = \mathbb{E}_{q}[f^2(x)\omega^2(x)] - I^2$.
Since the last term is independent of $q$, we can ignore it. Using Jensen's inequality, we have the following lower bound:
$$\mathbb{E}_{q}[f^2(x)\omega^2(x)]
\geqslant (\mathbb{E}_{q}[|f(x)\omega(x)|])^2 = \left(\int |f(x)|p(x)dx\right)^2$$
The lower bound is obtained when using the optimal importance distribution:
$q^*(x) = {|f(x)|p(x)}/{\int |f(x')|p(x')dx'}.$
Interestingly,  to estimate an integral,  it is more efficient to sample from $q(x) \propto |f(x)|p(x)$ than to sample from $p(x)$.  Of course, if  $f(x)$ is unknown, it is best to sample from $q(x) = p(x)$ \cite{murphy2012machine}.

\section{Positive Memory Retention}
This section contains our main contribution, the positive memory retention method.
The reuse of past positive trajectories in policy learning is enabled by several contributions: 
first, sample efficiency is improved by concentration of past positive trajectories; 
secondly, the sampling bias is corrected by importance sampling in policy gradients; 
thirdly, the stability is ensured by introducing bounds on the important weights; 
fourthly, the variance is reduced by probability updating of the proposal distribution; 
finally, the stability is improved by policy search via early stopping.



\textbf{Positive Memory Matters:}
In human memory retention, focusing on \textit{rewarded} events has been discovered to be a preferred strategy in the post-learning phase happening in the hippocampus area of the brain \cite{gruber2016post}. We believe that this fact also intuitively applies to RL since non-rewarded trajectories do not contribute directly to the estimated gradient to increase the expected return, $\nabla_{\theta} J(\theta)$, since $R(\tau)$ is zero.

In a more general case, consider the policy updating with a baseline function, e.g.\ $0<b<1$. The gradient of non-rewarded trajectories is opposite to the direction of the gradient of the current policy, $\nabla_{\theta}\mathrm{log} \pi_{\theta} (a_t|s_t)$, because $R(\tau) - b < 0$. This means that the weights are changed in a way to depress the current policy $\pi_{\theta}$, which is not necessarily equivalent to maximizing the expected return. However, it might be helpful, in a way to encourage exploration. 

The \emph{high efficiency} of positive memory retention can also be derived from importance sampling.
Consider that the expectation we want to estimate is the expected return $\mathbb{E}[R(\tau)]$, so $R(\tau)$ assumes the role of $f(x)$ in Section~\ref{sec:background}.
The proposal distribution $q$ should thus be of the form $q(\tau) \propto R(\tau)p(\tau)$, thus we only need to consider successful memory trajectories.




\textbf{Policy Gradient with Importance Sampling:} 
However, memory trajectories cannot be directly applied to policy gradient methods. The main reason is that the training requires  trajectories from the target policy
$p(\tau | \pi_{\theta} )$ with the current parameter vector $\theta$, whereas the memory trajectories were generated by $ q(\tau | \pi_{\theta'})= p(\tau | \pi_{\theta'} )$  with a different  parameter vector $\theta'$. We again can use the concept of IS and obtain 
\begin{equation*}
\hat{J}(\theta) = \frac{1}{n} \sum_{i=1}^{n} \frac{p(\tau^{(i)} | \pi_{\theta} )}{q(\tau^{(i)} | \pi_{\theta'})}R(\tau^{(i)}), \ \mathrm{with}\ \tau^{(i)} \sim q
\end{equation*}
where $n$ is the number of trajectories used to estimate the expected return $J(\theta)$ \cite{jie2010connection}.
In the equation above, we assume $q(\tau) = 0 \Rightarrow p(\tau) = 0$. This is readily true, since each action is sampled from the defined action space $\mathcal{A}$. 
The importance weights are evaluated using:
\begin{equation*}
\omega(\tau^{(i)}) = \frac{p(\tau^{(i)} | \pi_{\theta} )}{q(\tau^{(i)} | \pi_{\theta'})} = \frac{\prod_{t=1}^{T}\pi_{\theta(a_t | s_t)}}{\prod_{t=1}^{T}\pi_{\theta'(a_t | s_t)}}
\end{equation*}
where $\prod_{t=1}^{T}\pi_{\theta(a_t | s_t)}$ needs to be calculated from the target policy, and $\prod_{t=1}^{T}\pi_{\theta'(a_t | s_t)}$ has already been calculated from 
the behavior policy.
Finally, the importance weighted policy gradient is:
\begin{equation}
\nabla_{\theta} \hat{J}(\theta) = \nabla_{\theta} \mathbb{E}_q \left[\omega(\tau) (R^{}(\tau) - b) \mathrm{log} \pi_{\theta} (a_t|s_t) \right].
\label{eq:is_reinforce}
\end{equation} 

\textbf{Bounded Importance Weight Proposal:} 
This estimator is unbiased, but it suffers from very high variances because it involves a product of a series of unbounded importance weights. To prevent the importance weight from ``exploding'', the goal is now to select only samples that are \emph{not far} from the target policy.
 
%
To evaluate the distance, we use a symmetric version of the KL-divergence, i.e.\ the Jensen-Shannon divergence \cite{lin1991divergence}:
\begin{equation*}
\mathrm{JS}(p, q)\\
= 0.5 \mathbb{KL}(p \parallel 0.5(p+q)) 
+ 0.5 \mathbb{KL}(q \parallel 0.5(p+q)).
\end{equation*}
We now derive a formulation of the JS-divergence, as a distance metric, which is related to the importance weight $\omega$:
\begin{align*}
\mathrm{JS}(p, q)
&\approx 0.5 \sum_{k=1}^{K}p_k \mathrm{log}\frac{2p_k}{p_k+q_k} + 0.5 \sum_{k=1}^{K}q_k \mathrm{log}\frac{2q_k}{p_k+q_k} \\
&= 0.5 \sum_{k=1}^{K}p_k \mathrm{log}\frac{2}{1+\omega_k} + 0.5 \sum_{k=1}^{K}q_k \mathrm{log}\frac{2}{\frac{1}{\omega_k}+1} 
\end{align*}
We can see that the distance between the proposal distribution $q$ and the optimal solution $p$ depends on both $\omega_k$ and $1/\omega_k$. To limit the variance of the importance sampling, we limit the importance weight as $\omega_k \leq \omega_{max}$ and its inverse as $1/\omega_k \leq \omega_{max}$. Subsequently, we define a trust region of importance weights, $\omega_k \in [1/\omega_{max}, \omega_{max}]$ and only use trajectories whose importance weights fall within this range.


Essentially, we use the importance weight $\omega$ as a value to select high quality trajectories, filtering out those that deviate far from the current policy. In this way, we shape the proposal distribution into a safe one. The bound $\omega_{max}$ of the distribution can be selected by observing the learning curve during  training. 


\textbf{Probability Updating:}
Another way to reduce the variance is to \emph{adapt} the proposal distribution, $q(x)$, to make it as close as possible to $p(x)$. 
After updating the target policy with Equation \ref{eq:is_reinforce}, we use the current target policy as a behavior policy for retention in the future. 
In this way, the memory buffer is being continuously updated and the proposal distribution is also updated.

\textbf{Policy Search via Early Stopping:}
In order to make the best use of the memory, the learning process is verified using a group of validation samples. During the training process, the model remembers the positive trajectories within the current epoch for later retention. During the retention phase, the model first goes through the memory buffer, and updates the model using Equation \ref{eq:is_reinforce} with the bounded importance weight proposal. After each iteration, the model verifies the learned policy on a validation set. If the policy becomes better than the previous policy, then it is saved. If the policy has not been improved for a limited number of iterations $n_{max}$, then memory retention is stopped and the training of the model with REINFORCE continues. This mechanism helps the model make the best use of the past training samples and makes the learning more stable.  

\textbf{Complete Training Algorithm:}
\begin{algorithm}[t]
\caption{Positive Memory Retention (PMR)}\label{algo:complete}
\begin{algorithmic}[1] 
\Require{pretrained RNN language model $\pi_{\theta}$}
	\For{iteration in range(max iterations) }
	\For{$t=1\ \text{to}\ T$}
	\State{$(a_{t}, p_{t}) = \pi_{\theta}(x,y_{t-1})$, $y_{t} = (y_{t-1}, a_{t})$} 
	\EndFor
	\State{$r = \mathrm{R}(y) \leftarrow \mathrm{Environment}$}
	\For{$t =T\  \text{to}\  1$}
	\State{$\theta = \theta + \alpha (r - b) \nabla_{\theta}\mathrm{log} \pi_{\theta} (a_t|(x,y_{t-1}))$}
	\EndFor
	\If {$r>0$}
	\State{$\mathrm{memory} \leftarrow \tau = (x,y,p,r)$}
	\EndIf
	\EndFor
	\State{validating($\pi_{\theta}$), $\theta' = \theta$} 
	\For{trajectory $\tau$ in memory} \# memory retention
	\State{$x,y,p,r \leftarrow \tau$}
	\For{$t=1\ \text{to}\ T$}
	\State{$q_{t}=p_{t}$, $(a_{t}, p_{t})=\pi_{\theta}(x,y_{t-1})$}
	\State{$\text{memory} \leftarrow p_{t}$ \# probability updating}
	\EndFor
	\State{$\omega={\prod_{t=1}^{T}p_{t}}/{\prod_{t=1}^{T}q_{t}}$}
	\If{$\omega \in [1/\omega_{max},\omega_{max}]$}  
	\For{$t =T\  \text{to}\  1$}
	\State{$\theta = \theta + \alpha \omega  (r - b) \nabla_{\theta}\mathrm{log} \pi_{\theta} (a_t|(x,y_{t-1}))$}
	\EndFor
	\EndIf
	\EndFor
	\State{validating($\pi_{\theta}$)} 
	\If{no improvement for $n_{max}$ iterations}
	\State{$\theta = \theta'$} 
	\EndIf
	\If{improvement}
	\State{$\theta' = \theta$} 
	\EndIf
\end{algorithmic}
\end{algorithm}
We summarize the complete method in Algorithm \ref{algo:complete}.
Note that, before training the language model with RL, we pretrained the model in a supervised fashion for a kickstart policy.

\section{Experiments}
We conduct two sets of experiments to verify the proposed method. 
To highlight the methods' ability to boost sample-efficiency, we first create and experiment with a synthetic dataset \footnote{https://github.com/ruizhaogit/MNIST-GuessNumber}.
We then show that the method also works well on a real-word benchmark, GuessWhat?!\  \cite{guesswhat_game}. 

\subsection{MNIST GuessNumber Dataset}
\textbf{Experiment setting:} We create a synthetic dataset, named MNIST GuessNumber, which is designed for quick testing and analysis of RL methods in the task of visual-grounded goal-oriented dialog systems. The creation of the dataset is inspired by \cite{seo2017visual}. Each image in MNIST GuessNumber contains a $3\times3$ grid of MNIST digits and each MNIST digit in the grid has four randomly sampled attributes, i.e.\ $\mathrm{color}=\{ \mathrm{red},\ \mathrm{blue},\ \mathrm{green},\ \mathrm{purple},\ \mathrm{brown}\}$, $\mathrm{bgcolor}=\{ \mathrm{cyan},\ \mathrm{yellow},\ \mathrm{white},\ \mathrm{silver},\ \mathrm{salmon}\}$, $\mathrm{number}=\{ x|0\leq x \leq 9 \}$ and $\mathrm{style} = \{\mathrm{flat}, \mathrm{stroke}\}$, as illustrated in Figure \ref{fig:MNIST-GuessNumber-example-0}.
 
Given the generated image from MNIST GuessNumber, we automatically generate questions and answers about a set of the digits in the grid that focus on one of the four attributes. 
During question generation, the target subset for a question is selected based on the previous target subset referred by the previous QA, as shown in Figure \ref{fig:MNIST-GuessNumber-example-0}. 
For answer generation, we generate a yes/no answer based on whether the questioned attribute matches with the target digit.
The QA is repeated until there is only one digit in the target subset. 
We generated 30\,K, 10\,K, and 10\,K images for training, validating, and testing, respectively, and one successful game for each unique image. 
In each grid image, there are nine cells. Each cell contains four attributes, including the color, bgcolor, number, and the style.

\textbf{Model details and pretraining:}
There are three roles in this number guessing game, a questioner, an answerer, and a guesser.
The word and the image embeddings are trained end-to-end using lookup table layers.
The questioner model that we used is a long short-term memory (LSTM) \cite{hochreiter1997long} of 256 units conditioned on a given image.
The answerer model takes in the question along with the target digit and outputs a yes/no answer. 
The answer model is based on an LSTM with 64 units.
The last one is the guesser model. 
The guesser uses an LSTM with 64 units to process the whole dialog, and compares it with each digit using a dot product on their respective latent representations. The prediction is the most similar digit.
We train all three models for 30 epochs using the maximum likelihood criterion.
The pretrained models obtain a game success rate of 63.09\% on the test split with a maximum of four rounds of QA.

\textbf{Reinforced training:}
After pretraining, we keep the answerer and the guesser fixed and train the questioner model with RL.
Given the unique images in the training set, for each game, the answerer randomly picks a digit in the image as the target and lets the questioner ask.
The baseline method is the REINFORCE. 
For positive memory retention, we set the parameter $\omega_{max}=10$, and the early stopping threshed $n_{max}=2$.
The $\omega_{max}$ is selected on the validation set. An extensive evaluation of the impact of $\omega_{max}$ is shown in Table \ref{tab:ablation} lower part.
When we use $\omega_{max}=10$, we observe that about 65\% of the positive trajectories are used for weight updating.
So, $\omega_{max}$ can be considered as a trade-off between sample reuse ratio and the variance introduced by importance sampling.
The early stopping threshed $n_{max}$ is a trade-off between sample-efficiency and computational power.
Our implementation \footnote{https://github.com/ruizhaogit/PositiveMemoryRetention} uses Torch7 \cite{torch}. 

\textbf{Results:}
\begin{figure*}
    \centering
    \subfloat{{\includegraphics[width=7cm]{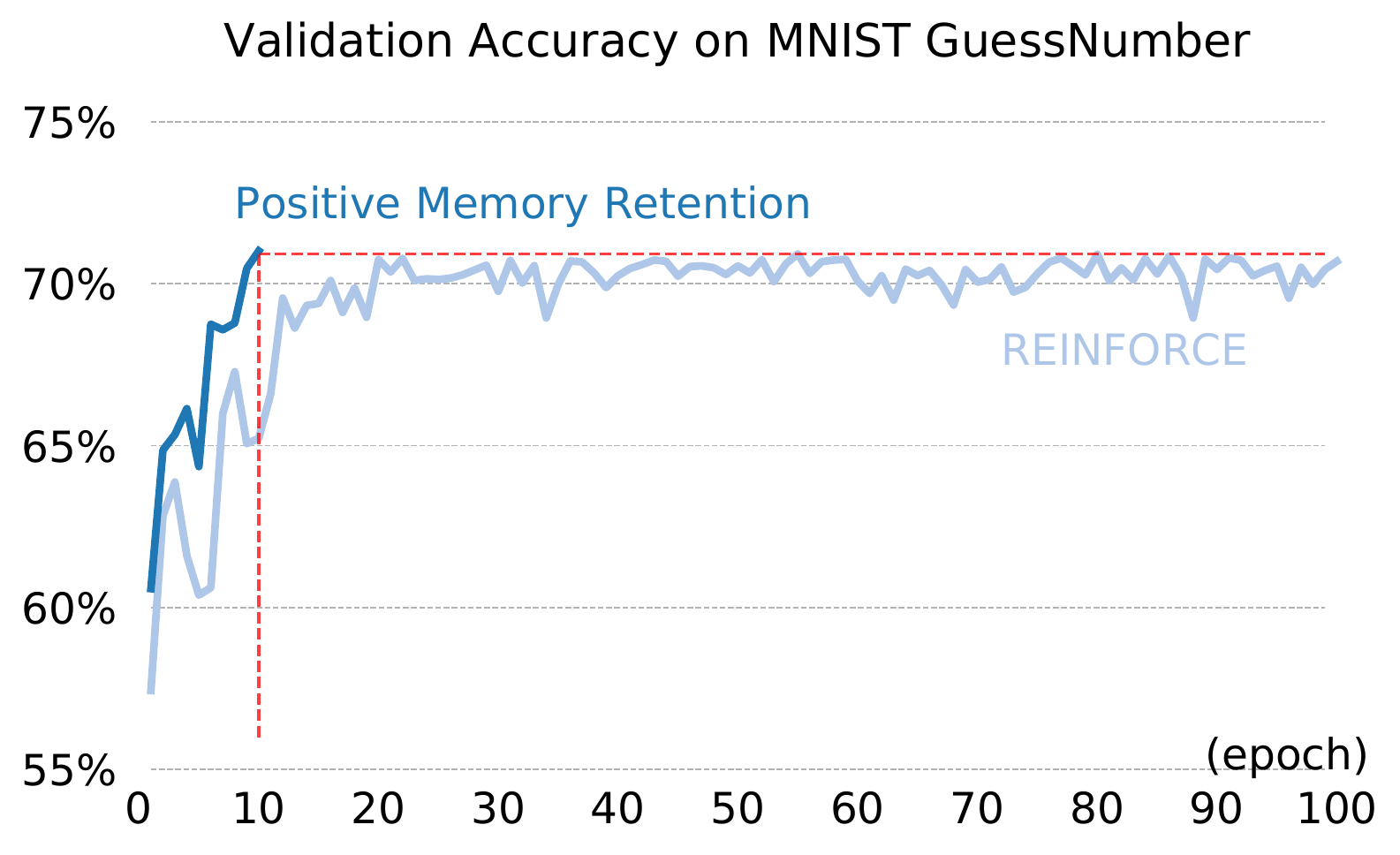} }}
    \qquad
    \qquad
    \subfloat{{\includegraphics[width=7cm]{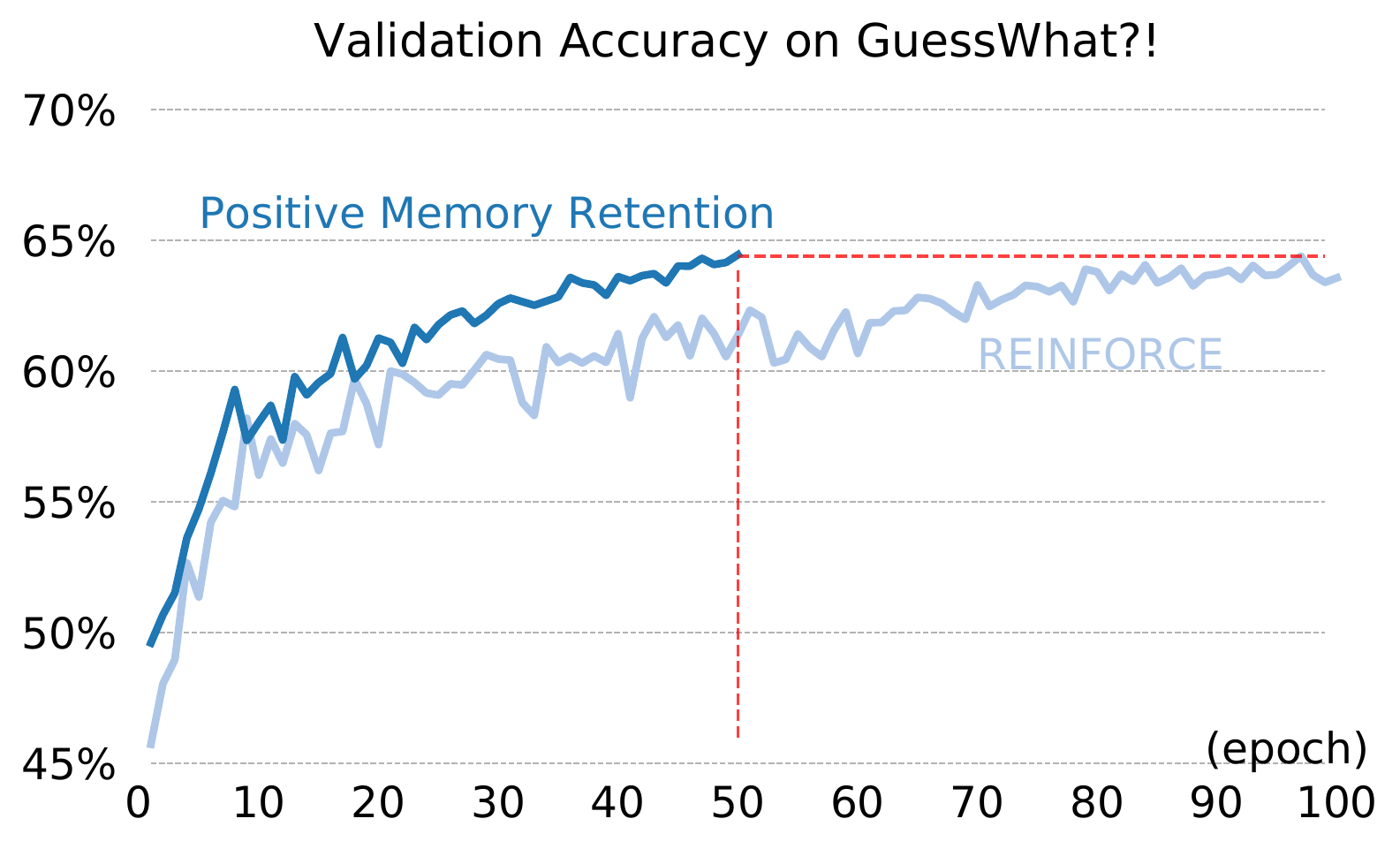} }}
    \caption{\textbf{Experimental results:}
    The left figure shows that at the 10\textsuperscript{th} epoch, the model trained with positive memory retention obtains about 70\% accuracy on the validation set, which is is equivalent to the same model trained with REINFORCE for 100 epochs, which obtains 69.92\% accuracy.
    The right figure shows that the model trained with REINFORCE for 100 epochs, obtains 63.39\% accuracy on the validation split. At the 50\textsuperscript{th} epoch, the same model trained with our proposed method has a better result, 63.44\%.}%
    \label{fig:plot}
\end{figure*}
From Figure \ref{fig:plot} left, we can see that at the 10\textsuperscript{th} epoch, the model trained with positive memory retention, obtains about 70\% accuracy on the validation set, which is comparable with the same model trained with REINFORCE for 100 epochs with an accuracy of  69.92\%. 
After training, we evaluated the best model on the test set.
REINFORCE and positive memory retention obtain 69.86\% and 70.27\% accuracy on the test split, respectively.  
However, the memory positive attention only needs one-tenth of the training samples.
The experiment on MNIST GuessNumber shows that the sample-efficiency has been improved by a factor of 10.

\textbf{Ablation tests:}
\begin{table}[t]
\centering
\caption{\textbf{Ablation tests on MNIST GuessNumber:} Notations: RF denotes the REINFORCE; IS is importance sampling; PM means using positive memory only, otherwise all memory; UB denotes the upper bound $\omega_{max}$; LB represents the lower bound $1/\omega_{max}$; PB is the probability updating trick: ES means the early stopping within epochs, if use ES, then the early stopping threshold is 2, otherwise train for 3 iterations; (\%) is the accuracy on the test split using the best model selected via validation split during 10 epochs of training. 
The \textbf{upper part} above the middle horizontal line shows the ablation test of different components in the positive memory retention. Here, $\omega_{max}=10$, and (\#\,1) is the performance of the kickstart policy after supervised training. Note that (\#\,3) and (\#\,4) \emph{diverges} quickly, which means that the testing accuracies are lower than 20.0\% after 10 epochs of training. UB makes the stable training of RF+IS possible, shown in (\#\,5). PM, LB, PB, ES, contribute 0.76\%, 1.18\%, 0.73\%, and 1.54\%, respectively.
The \textbf{lower part} below the middle horizontal line shows the extensive evaluation regarding to the upper bound parameter $\omega_{max}$.
}

\begin{tabular}{
p{0.3cm} p{0.3cm} p{0.3cm} p{0.3cm} p{0.3cm} p{0.3cm} p{0.3cm} p{0.3cm} p{1cm} }
  \# & RF & IS & PM & UB & LB & PB & ES & \ (\%) \\ \toprule
  \\[-.8em]
  1  & -- & -- & -- & -- & -- & -- & -- & 63.09 	\\
  2  & \c & -- & -- & -- & -- & -- & -- & 65.40 	\\
  3  & \c & \c & -- & -- & -- & -- & -- & < 20. 	\\
  4  & \c & \c & \c & -- & -- & -- & -- & < 20.	\\ 
  5  & \c & \c & -- & \c & -- & -- & -- & 66.06	\\
  6  & \c & \c & -- & \c & \c & -- & -- & 67.24	\\
  7  & \c & \c & \c & \c & \c & -- & -- & 68.00	\\
  8  & \c & \c & \c & \c & \c & \c & -- & 68.73	\\
  9  & \c & \c & \c & \c & \c & \c & \c & \textbf{70.27}	\\ \midrule
 10  & \c & \c & \c & 1 & \c & \c & \c & 66.24 \\
 11  & \c & \c & \c & 5 & \c & \c & \c & 65.69 \\
 12  & \c & \c & \c & 10 & \c & \c & \c & \textbf{70.27} \\
 13  & \c & \c & \c & 20 & \c & \c & \c & 69.38 \\
 14  & \c & \c & \c & 30 & \c & \c & \c & 69.09 \\
 15  & \c & \c & \c & 100 & \c & \c & \c & 65.39 \\ \bottomrule
\end{tabular}
\label{tab:ablation}
\vspace{-1em}
\end{table}
A summary of the ablation tests is shown in Table \ref{tab:ablation}. We can see that the bounded importance weight proposal is critical for policy gradient with importance sampling. Without this component, the model \emph{diverges} quickly, shown in Table \ref{tab:ablation} (\#\,3-4). 
The other proposed innovations all improve model performance as well, as shown in Table \ref{tab:ablation} (\#\,5-9). 
One can also see that the choice of the bound parameter $\omega_{max}$ has a major influence on the performance.

\subsection{GuessWhat?!\ Game}

\textbf{Experiment settings:}
In the GuessWhat?!\ dataset \cite{guesswhat_game} the dialogs are collected using Amazon Mechanical Turk with respect to MS-COCO \cite{lin2014microsoft} images. 
Each game is composed of an image, a target object in the image, the spatial information of the objects, the category of the objects, and the QA-dialogs. Unlike the MNIST GuessNumber, the questions in the training set are in free form text. The answers are still in the yes/no form. This dataset is more challenging due to its large vocabulary size (5\,K), and long dialog sequences.
Due to the large action space and the extremely delayed reward signals, the importance sampling estimates have very large variances. 
In our experiment, when we first attempted to retain with all past trajectories, and without the weight bound, the model \emph{diverged} quickly, as in the MNIST GuessNumber experiments, see Table \ref{tab:ablation} (\#\,3). Our proposed method reduces the variance and makes the sample reuse possible for real-word sceneries.

\textbf{Model details and pretraining:}
Each game in the GuessWhat?!\ contains three roles, a questioner, an answerer, and a guesser. As our aim is to compare the sample-efficiency of our proposed model with other strong baselines, we use the same model structure as was used in \cite{end_to_end_gw}.
The questioner model is a one layer LSTM with 512 units and conditioned on the VGG16-CNN-FC8 \cite{simonyan2014very} features of the image. 
The answerer model deploys an LSTM with 512 units to process the question along with spatial and categorical information of the target object. The guesser model uses an LSTM to process the whole dialog and can consider all the spatial and categorical information of the objects in the image.
The guesser compares the similarities between the dialog representation and each object representation with a dot product, and then takes the guess. All these three models are pretrained with MLE for 30 epochs for a kickstart policy. We reproduced the paper's experimental results using Torch7 \cite{torch}, and obtained 41.41\% accuracy on the test split after supervised training.

\textbf{Reinforced training:}
With the pretrained models, we keep the answerer and the guesser fixed and train the questioner model. We train the model for 100 epochs, using REINFORCE with a learning rate $\alpha$ of 0.0001 and a running average as the baseline $b$. Our reimplementation using Torch obtains 62.61\% accuracy on the test split, about 2\% higher than their result of 60.3\% from the original implementation \cite{guesswhat_github}, due to some technical differences. We use our reimplementation as the REINFORCE baseline, in Figure \ref{fig:plot}, to eliminate the influence of these technical differences for a fair comparison. For positive memory retention, Algorithm \ref{algo:complete}, we use weight bound $\omega_{max} = 5$, so that $\omega \in [1/5, 5]$, to stabilize the training. We use the early stopping threshed in each epoch as $n_{max}=2$. We observe that with $\omega_{max} = 5$, about 85\% of the trajectories in the memory contribute to the model weight updates. This high ratio of reuse is also due to the probability updating mechanism, which bridges the gap between the behavior policies and the target policy.

\textbf{Results:}
From Figure \ref{fig:plot} right, we can see that the model trained with REINFORCE obtains 63.39\% accuracy on the validation split after training for 100 epochs. However, at the 50\textsuperscript{th} epoch, the same model trained with positive memory retention reaches 63.44\% validation accuracy. 
After training, we evaluated the best model on the test set.
REINFORCE and positive memory retention obtain 62.61\% and 63.17\% accuracy on the test split, respectively.
We can see that the proposed method provides state-of-the-art performance with double sample-efficiency on the GuessWhat?!\ dataset.

\section{Related Work}

\textbf{Goal-Oriented Dialogs:}
Recently, researchers started exploring intensively deep RL for goal-oriented dialogs \cite{bordes2016learning,end_to_end_gw,visdial_rl,williams2016end,dhingra2016end,lipton2016efficient,li2016deep,li2017end,rudnicky1999agenda,singh2000reinforcement,lemon2006hierarchical,zhao2016towards,zhao2018improving,zhao2018learning,asadi2016sample,weisz2018sample,su2017sample,chen2017agent}, focusing on learning to achieve a goal via dialog.
Bordes and Weston \cite{bordes2016learning} pointed out that the recent success in chit-chat dialogs may not carry over to goal-oriented settings. 
Strub et al.\ \cite{end_to_end_gw} and Das et al.\ \cite{visdial_rl} conduct the task-oriented conversation over image guessing games. 
In \cite{end_to_end_gw}, the dialog aims at object discovery through a series of yes/no QAs. Policy gradient is used to improve performances of dialog agents in terms of task completion rate. 
Das et al.\ \cite{visdial_rl} use policy gradient to train two chat-bots to play image guessing games and show that they establish their own communication style. 
Both works use on-policy policy gradient methods. Sample-efficient on-off-policy learning methods, as developed in this paper, have not yet been explored in goal-oriented dialogs.

\textbf{Sample-Efficient RL:}
While the goal-oriented dialog using RL is a recent research direction, control tasks via RL have been studied extensively and importance sampling based actor-critic methods have been known to be beneficial for sample-efficiency \cite{jie2010connection,degris2012off,munos2016safe,wang2016sample,gruslys2017reactor,espeholt2018impala}. However, the control tasks are inherently different from dialog tasks in the aspect of action space. For example, in Atari games, the agents normally have less than 20 actions to explore; in contrast, the action space, i.e.\ the vocabulary, contains thousands of words in dialogs. Moreover, the reward of a dialog is only available at the end, which is much more sparse and delayed than in Atari games. In these games, there are intermediate rewards prior to the game ending. 
The long trajectories in dialog tasks make the often observed problem of exploding importance weights even more extreme.
Even if an explosion does not occur, the variance of the importance sampling increases.
To overcome these challenges, new solutions must be introduced.

\textbf{Memory Retention:} 
The use of positive memory retention is inspired by recent neuroscience research, which concludes that the brain prioritizes those high-reward memories, which might be the most important for obtaining future rewards \cite{gruber2016post}. 
Tresp et al.\ \cite{tresp2015learning} argue that the brain's memory functions might inspire technical solutions requiring memory traces.
Biologically-inspired experience replay \cite{mnih2015human}, was used to stabilize the training process in RL and and thereby was quite successful. 
These papers used Q-learning, which is an off-policy method that is able to use the past trajectories directly.
However, on-policy policy gradients cannot reuse past samples directly \cite{jie2010connection,degris2012off}.
The main contribution of this paper is that we show that our extensions permit an efficient reuse of past samples in on-policy policy gradients methods. These extensions also work well in dialog settings, which are challenging due to the sparse reward and the large action space.


\section{Conclusion}
We proposed a novel positive memory retention mechanism for improving sample-efficiency in dialog policy learning, using past positive trajectories and low-variance importance sampling estimates. 
The model reuses past positive samples as behavior policies, samples from a bounded importance weight proposal, and updates the target policy with an importance weight correction. 
We tested on both synthetic and real-word datasets and illustrated dramatically improved sample-efficiency.
We demonstrate that policy gradient can successfully be trained using past trajectories in dialog tasks.


\bibliographystyle{IEEEbib}
\bibliography{reference}

\end{document}